\documentclass[letterpaper, 10 pt, conference]{IEEEtran}
\usepackage{graphics} 
\usepackage{epsfig} 
\usepackage{mathptmx} 
\usepackage{times} 
\usepackage{cite}
\usepackage{amsmath} 
\usepackage{amssymb}  
\usepackage{graphicx}
\usepackage{mwe}
\usepackage{lipsum}%
\usepackage{booktabs}
\usepackage[caption=false]{subfig}
\usepackage{tabularx}
\usepackage{multirow}
\usepackage{gensymb}

\usepackage{footnote}
\usepackage[para,online,flushleft]{threeparttable}

\makesavenoteenv{tabular}
\makesavenoteenv{table}

\begin{document}
	
	\title{\LARGE \bf
		Domain Adaptation for Vehicle Detection from Bird's Eye View LiDAR Point Cloud Data}
	 \author{\IEEEauthorblockN{Khaled Saleh, Ahmed Abobakr, Mohammed Attia, Julie Iskander, Darius Nahavandi and
	 		Mohammed Hossny\\~\\ 
	 		Institute for Intelligent Systems Research and Innovation (IISRI)\\
	 		Deakin University\\~\\ k.aboufarw@deakin.edu.au
	 		}}

	\maketitle
	
	\begin{abstract}
Point cloud data from 3D LiDAR sensors are one
of the most crucial sensor modalities for versatile safety-critical
applications such as self-driving vehicles. Since the annotations
of point cloud data is an expensive and time-consuming process,
therefore recently the utilisation of simulated environments and
3D LiDAR sensors for this task started to get some popularity. With simulated sensors and environments, the process
for obtaining an annotated synthetic point cloud data became
much easier. However, the generated synthetic point cloud data
are still missing the artefacts usually exist in point cloud data
from real 3D LiDAR sensors. As a result, the performance
of the trained models on this data for perception tasks when
tested on real point cloud data is degraded due to the domain
shift between simulated and real environments. Thus, in this
work, we are proposing a domain adaptation framework for
bridging this gap between synthetic and real point cloud data.
Our proposed framework is based on the deep cycle-consistent
generative adversarial networks (CycleGAN) architecture. We
have evaluated the performance of our proposed framework
on the task of vehicle detection from a bird's eye view (BEV)
point cloud images coming from real 3D LiDAR sensors. The framework has shown competitive results with an improvement of more than 7\% in average precision score over other baseline approaches when tested on real BEV point cloud images. 
		
	\end{abstract}

\section{Introduction}
Recently, deep learning-based techniques such as convolution neural networks (ConvNets) have been achieving state-of-the-art results in many computer vision tasks such: object identification~\cite{yolov3}, scene understanding~\cite{saleh2018end,saleh2018local}, and human action recognition~\cite{saleh2018cyclist,bakr2019fall,saleh2018long}. However, these techniques require a handful amount of labelled data for training them which is both time-consuming and cumbersome to get for many tasks. Thus, the utilisation of synthetic data for training such techniques got some momentum over the past few years~\cite{saleh2017cyclist,saleh2018effective}. With synthetic data, the process for obtaining ground-truth labels becomes much easier and automated most of the time. However, still, the utilisation of synthetic data is not entirely reliable because of its limitations when it comes to the generalisation to real data. 

\begin{figure}[t]
 \centering
	\includegraphics[width=.49\columnwidth]{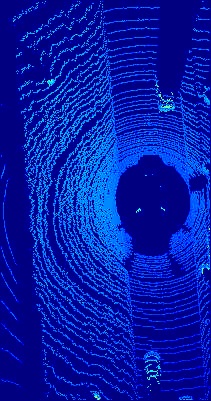}
	\includegraphics[width=.49\columnwidth]{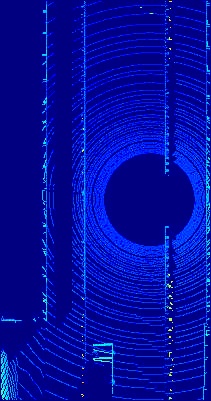}
		\caption{Sample of BEV images of real point cloud data (left) from a real Velodyne 3d LiDAR from KITTI dataset~\cite{Geiger2012CVPR} and a synthetic point cloud data (right) from a simulated 3D LiDAR sensor from MDLS dataset~\cite{yoon2018mapless}.} 
	\label{fig:syn_velo}
\end{figure}

\smallbreak
In safety-critical applications such as a self-driving vehicle, one of the main sensors that are currently crucial for its development is the 3D LiDAR (Light Detection And Ranging) sensor. 3D LiDAR sensors can reliably provide 360$\degree$ point cloud  in traffic environment with coverage distance up to 200 meters ahead across different weather and lighting conditions. Thus, a number of deep-learning based techniques have recently been utilising its point cloud for many perception tasks for self-driving vehicles~\cite{yoon2018mapless,wu2018squeezesegv2}. The reason that the number of deep-learning techniques that rely on point-cloud data is not as much as the ones rely on visual data is the scarcity of labelled point cloud data. The labelling procedure for point cloud data is more complicated than visual data especially for tasks such as 3D object detection and per-point semantic segmentation. Thus, the usage of synthetic data has been explored, similar to the visual data modality data~\cite{saleh2017cyclist, yoon2018mapless}. However, the generalisation to real-point cloud data was rather limited due to the perfectness of the synthetic point cloud data (shown in Fig.~\ref{fig:syn_velo}, right) which is missing the artefacts usually exist in point cloud data from real 3D LiDAR sensors (shown in Fig.~\ref{fig:syn_velo}, left). These artefacts are such as the variability of the LiDAR beams intensities or the motion distortion as a result of the motion of the 3D LiDAR. 

\smallbreak
Domain adaptation (DA) is one of the machine learning (ML) techniques that have been recently explored to bridge the aforementioned gaps between synthetic and real data domains~\cite{wang2018deep}. In DA, the goal is to learn from one data distribution (referred to as the source domain) a perfect model on a different data distribution (referred to as the target domain). In traffic environments, DA has recently shown promising results for image translation between different domain pairs such as night/day, synthetic/real images and RGB/thermal images~\cite{zhu2017unpaired}. Since most of the previous DA techniques are based on 2D deep ConvNet architectures, thus their application on 3D point cloud data from 3D LiDAR sensors is not a straight forward task.


\smallbreak
On the other hand, the recent deep-learning based techniques that have been applied on perception tasks using 3D point cloud data, they managed to find a way to adopt the same 2D ConvNet architectures to work on the 3D point cloud data. One of the most common techniques was to project a top-down bird's eye view (BEV) of the point cloud data on a 2D plane (ie. ground). The representation of the 3D LiDAR point cloud data as a BEV was shown to be effective in many perception tasks for self-driving vehicles such as 3D object detection~\cite{li20173d}, road detection~\cite{caltagirone2017fast} and per-point semantic segmentation~\cite{dewan2017deep}. 
\smallbreak
To this end, in this work, we will be proposing a DA approach for vehicle detection in real point cloud data from 3D LiDAR sensors represented as BEV images. The proposed DA approach will be a deep learning-based approach based on deep generative adversarial networks (GANs)~\cite{zhu2017unpaired}. For the vehicle detection task, it will be based on state-of-the-art deep object detection architecture YOLOv3~\cite{yolov3}. The rest of the paper is organised as follows. In Section~\ref{related}, a brief introduction to the different DA approaches with emphasis on deep learning based approaches will be reviewed in addition to a quick review on GANs. Section~\ref{method}, the methodology we followed for our proposed DA approach will be discussed thoroughly. Experiments and results are discussed in Section~\ref{exper}. Finally, Section~\ref{concl} concludes.

\section{Related Work}\label{related}
Commonly, there are two ways to achieve DA either by directly translating one domain to the other or by obtaining a common-ground intermediate pseudo-domain between the two domains. In the following, firstly a quick review of the work related to the DA approaches will be provided specifically the approaches based on the direct translation between domains. Then, a brief summary of the DA work between simulated and real domains done in the context of traffic environments will be discussed.

\subsection{Adversarial Domain Adaptation}\label{da}
Historically, most of the work done on DA has been relying on the transformation between source and target domains based on linear representations~\cite{blitzer2006domain,germain2013pac}. Until the emergence of the recent set of techniques based on non-linear transformation representations via neural networks~\cite{ganin2016domain,tzeng2017adversarial}, which have achieved state-of-the-art results in a number of DA benchmarks~\cite{netzer2011,lecun1998gradient}. One of the most commonly non-linear-based representations DA approaches is the adversarial domain adaptation (ADA) approach~\cite{ganin2016domain}. ADA was inspired by the work done by Goodfellow et al.~\cite{goodfellow2014generative} on generative adversarial networks (GANs). In GANs, there are two deep neural networks trained simultaneously, namely a ``generator'' network and a ``discriminator'' network. The generator network, as the name implies, it generates new data instances using a uniform distribution, on the other hand, the discriminator network tries to decide whether or not this newly generated data instance has the same distribution as the training dataset distribution. Similarly, in ADA, it has the same two networks, where the generator network, generates instances from the source domain distribution to transform it into the target domain distribution. Whereas, the discriminator network tries to differentiate between the instances outputted from the actual target domain distribution and the ones generated from the generator network. Thus, this architecture is often referred to in the literature as the ``conditional GAN''. One of the most recently successful ADA architectures is the Cycle-Consistent GAN (CycleGAN)~\cite{zhu2017unpaired} architecture. In CycleGAN, it is essentially comprised of two conditional GAN networks. The first network works on the transformation from the source domain ($S$) to the target domain ($T$), $S \rightarrow T$, while the other one works on the transformation in the opposite direction, $ T \rightarrow S$. The additional contribution for CycleGAN architecture was the introduction of a new loss function they call it the cycle-consistency loss function. This new loss function assures that if the two conditional GANs networks are connected, they will produce the following identity mapping: $S \rightarrow T \rightarrow S$. 

\subsection{DA Between Synthetic and Real for Perception Tasks}\label{dgap}
In the context of traffic environments, a number of perceptions tasks has been utilising the DA approach to bridge the gap between real domains from physical sensors and synthetic domains from simulated sensors~\cite{zhu2017unpaired, atapour2018real,zhang2019synthetic}. It is worth noting that all of these works were only exploring one type of sensors which was cameras either RGB (monocular/stereo) or thermal. For example, in~\cite{zhu2017unpaired}, a number of DA between different domains were introduced based on the CycleGAN architecture. For instance, they addressed the semantic segmentation task between the day and night domains on unpaired visual images from multiple road-based datasets. Similarly, in~\cite{atapour2018real}, Atapour et al. trained a ConvNet model on synthetic depth and RGB images from the famous game GTA in order to estimate a synthetic monocular depth image. In the testing/inference phase, they took an input real RGB image from the KITTI dataset~\cite{menze2015object} and with the help of a CycleGAN architecture, they transformed the real RGB image into a synthetic GTA game like RGB image. Then, they passed the synthetic RGB image to their initial trained model to estimate a synthetic depth image. Eventually, they used the same CycleGAN network again to adopt the estimated depth image from the synthetic image domain to a real RGB image domain.

\smallbreak
On the other hand, in~\cite{zhang2019synthetic} Zhang et al. proposed deep-learning based approach for thermal infra-red object tracking. To overcome the scarcity of thermal images dataset, they utilised DA based on the CycleGAN architecture to transform images from visual domain to the thermal infra-red domain. 
	\begin{figure*}[t]
		\centering
		\includegraphics[height=12cm, keepaspectratio=true]{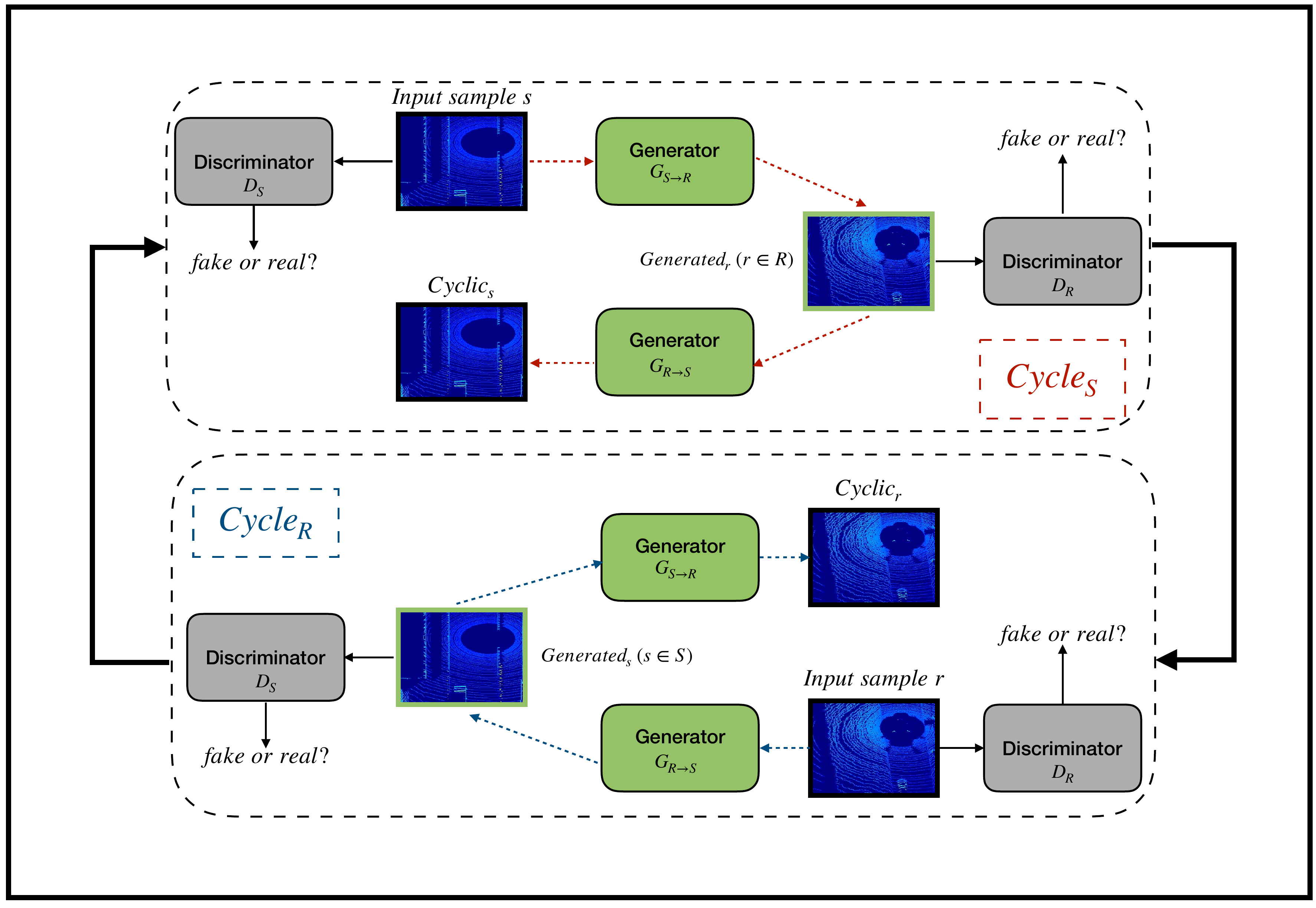}
		\caption{Proposed CycleGAN-based DA framework for the vehicle detection task in BEV point cloud images. The framework has two internal cycles, namely $Cycle_S$ and $Cycle_R$. In $Cycle_S$, the input sample $s$ of synthetic BEV point cloud image goes firstly through the generator $G_{S\rightarrow R}$ which its output is interrogated by the discriminator $D_R$. The generated  sample $r$ is then goes through the other generator $G_{R\rightarrow S}$ for reconstructed the original input $s$ sample. The same process goes for the second cycle $Cycle_S$.}
		\label{fig:frmwrk}
	\end{figure*}
\section{Proposed Methodology}\label{method}
The main focus of this work is to provide a framework for bridging the gap between real and synthetic point cloud data represented as BEV images for the vehicle detection task. That being said, the same framework can still be used for other perceptions tasks on point cloud data such as semantic segmentation or object tracking. In this section, we will first provide our formulation for the problem at hand. Then subsequently, we will break-down the building blocks of the proposed framework. 

\subsection{Problem Formulation}
ConvNet-based architectures for object detection from BEV point cloud data has been achieving state-of-the-art results in many benchmarks~\cite{li20173d}. However, with the available insufficient numbers of annotated BEV point cloud data for training such architectures, the trained models are still performing poorly especially in challenging scenarios. The utilisation of annotated synthetic BEV point cloud data from simulated traffic environments could be the key to increase the performance of such models. However, due to the domain shift between real and synthetic BEV point cloud data, the trained model on synthetic data is not necessarily guaranteed to generalise on the real data~\cite{wu2018squeezesegv2}. 


\smallbreak
Thus, in our formulation for the vehicle detection task from real BEV point cloud data, we are proposing a framework for DA between synthetic BEV point cloud data and real BEV point cloud data. In the first stage of our framework, we train a CycleGAN model between unpaired synthetic BEV point cloud data and real BEV point cloud data. The trained model, in returns, learns a transformation from synthetic BEV point cloud data to real BEV point cloud data and vice versa. As a result, given any annotated synthetic BEV point cloud dataset with vehicles, the trained CycleGAN model will transform that dataset to an annotated real-like BEV point cloud data. Finally, using the transformed dataset, we could train another ConvNet-based model for the vehicle detection task in real BEV point cloud data.  

\subsection{Deep Unsupervised DA via Cycle-Consistent GANs} \label{cycGAN}
As we earlier mentioned in Section~\ref{dgap}, the CycleGAN architecture has recently shown promising results in a number of DA tasks between real and synthetic visual domains. Thus, in this work, we will be exploring the CycleGAN architecture for the task of DA between real BEV point cloud domain and synthetic BEV point cloud domain. One of the advantages of the CycleGAN architecture in the context of DA is it can learn transformation between source and target domains without any supervised one-to-one mapping between the two domains. This is beneficial for our task because it is almost impossible for us to have the same traffic scenario and environment captured in both real BEV point cloud data and synthetic BEV point cloud data. However, we can have a handful amount of BEV point cloud data from each domain separately that represent the distribution of that domain.  

\smallbreak
More formally, given our two domains $S, R$ of the synthetic and the real BEV point cloud data domains. Then, the objective of our adopted CycleGAN-based DA approach (shown in Fig.~\ref{fig:frmwrk}) is to map between the distributions $s \sim \mathbb{P}_{d}(s)$ and $r \sim \mathbb{P}_{d}(r)$ from the synthetic and the real BEV point cloud domains respectively. The proposed CycleGAN-based DA approach achieve this mapping via the two generators, $G_{S\rightarrow R}$ and $G_{R\rightarrow S}$ and the two discriminators $D_{S}$ and $D_{R}$. The generator $G_{S\rightarrow R}$ will try to map the input source synthetic BEV point cloud image to some target real BEV point cloud image. While the generator $G_{R\rightarrow S}$ is trying to map the generated BEV point cloud image from the real target domain back to its original source domain. The discriminator $D_{S}$, on the other hand, is trying to differentiate between a BEV point cloud image $s \in S$ and a generated BEV point cloud image from $G_{R\rightarrow S}$. Conversely, the discriminator $D_{R}$ will be trying to distinguish between a BEV point cloud image $r \in R$ and a generated BEV point cloud image from $G_{S\rightarrow R}$. The two generators networks are deep ConvNet models. 

\smallbreak
The main building blocks of them are three blocks, namely the encoder, the transformer and the decoder respectively. The encoder's job is to extract features on multiple levels progressively by down-sampling them from the input BEV point cloud image from both domains. The transformer, on the other hand, takes the extracted features vector encoder in the source domain and transform it into another feature vector in the opposite target domain. The decoder finally up-sample the transformed features vector back to the original shape and dimensionality as it was before going through the encoder. The architecture we used for that combination of encoder, transformer and decoder of our generator networks is based on the architecture proposed in~\cite{johnson2016perceptual}. The encoder in this architecture consists of two convolution layers, while the transformer consists of nine ResNet blocks and the decoder consists of two de-convolution/transposed convolution layers. The two discriminators architecture is a deep ConvNet model as well. They are based on the PatchGAN architecture from~\cite{isola2017image}, which consists of three consecutive convolution layers for feature extraction in patches and a final 1D-convolution layer for the decision whether its input BEV point cloud image is fake or not. 

\smallbreak
In order to train the proposed CycleGAN-based DA approach for our task, we will be utilising the adversarial loss for the two generators that we have discussed above along with their corresponding discriminators.  The first loss for the transformation from domain $S$ to domain $R$ is as follows:

\begin{equation}
\label{eq:first_loss}
\begin{split} 
\mathcal{L}_{adv_{S\rightarrow R}}  =\underset{G_{S\rightarrow R}}\min \underset{D_{R}} \max  \underset{r \sim \mathbb{P}_{d}(r)}{\mathbb{E}}[\log D_R(r)] + \\
 \underset{s \sim \mathbb{P}_{d}(s)}{\mathbb{E}}[\log(1-  D_R(G_{S\rightarrow R}(s)))] 
\end{split}
\end{equation}
where $S$ is the synthetic BEV point cloud data domain and  $\mathbb{P}_{d}(s)$ is its data distribution.

\smallbreak
Similarly, the second loss for the transformation from domain $R$ to domain $S$ is as follows:
\begin{equation}
\label{eq:second_loss}
\begin{split} 
\mathcal{L}_{adv_{R\rightarrow S}}  =\underset{G_{R\rightarrow S}}\min \underset{D_{S}} \max  \underset{s \sim \mathbb{P}_{d}(s)}{\mathbb{E}}[\log D_S(s)] + \\
\underset{r \sim \mathbb{P}_{d}(r)}{\mathbb{E}}[\log(1-  D_S(G_{R\rightarrow S}(r)))] 
\end{split}
\end{equation}

Additionally, in order to penalise the generators of the trained model to generate more realistic BEV point cloud data from each domain $S$ and $R$, the following third loss is added.

\begin{equation}
\label{eq:third_loss}
\begin{aligned} \mathcal{L}_{c y c} &=\left\|G_{R\rightarrow S}\left(G_{S\rightarrow R}(s)\right)-s\right\|_{1} \\ &+\left\|G_{S\rightarrow R}\left(G_{R\rightarrow S}(r)\right)-r\right\|_{1} \end{aligned}
\end{equation}
where $\mathcal{L}_{c y c}$ is the cycle-consistency loss which ensures the identity mapping of the each transformed sample BEV point cloud image back to its original source. 

\smallbreak
Given the three losses from Eq.~\ref{eq:first_loss},~\ref{eq:second_loss},~\ref{eq:third_loss}, the objective loss function for the proposed CycleGAN-based DA approach is as follows:

\begin{equation}
\label{eq:total_loss}
\mathcal{L}=\mathcal{L}_{adv_{S\rightarrow R}}+\mathcal{L}_{adv_{R\rightarrow S}}+\lambda \mathcal{L}_{c y c}
\end{equation}
where $\lambda$ is equal to 10 which was chosen empirically. 

\smallbreak
Finally, since the objective of training any deep ConvNet model is to minimise a certain loss function,  which in our case is the joint loss function in Eq.~\ref{eq:total_loss}. Thus, we will be using the Adam optimiser for minimising our objective joint loss function using a learning rate of 0.001.

\subsection{Vehicle Detection in BEV Point Cloud Data via YOLOv3}
For the vehicle detection task, we will be the adopting state-of-the-art single stage deep ConvNet architecture for object detection, You Only Look Once (YOLOv3) architecture. Internally, YOLOv3 relies on k-means clustering to have prior bounding boxes ``anchors'' of a potential region of interests (ROIs) in the input image which goes through a total of 53 convolution layers to extract features from them on 3 different scales. YOLOv3 in returns predicts the four coordinates for the bounding box, an objectness score for each bounding box, and class score for the object that the bounding box may contain. The four coordinates are predicted using a sigmoid function. The objectness score is predicted using a logistic regression which is set to 1 if the bounding box of one of the anchors overlaps with a ground truth bounding box. The class score of a bounding box is predicted via multinomial logistic classifiers which is better than the traditional soft-max classifier when it comes to multi-label classification task such as object detection. 

\smallbreak
More specifically, in our vehicle detection task from BEV point cloud images, we relied on the YOLOv3-416 derivative architecture, which as the name implies works on input images with a resolution of $416 H \times 416 W$. 

\section{Experiments}\label{exper}
In this section, we will firstly discuss the datasets we have used for training and validating our trained models. Secondly, the performance of our models will be quantitatively and qualitatively evaluated. 
\subsection{Datasets}
For the task of the DA between synthetic and real BEV point cloud images, we relied on two datasets. The first dataset is the recently released Motion-Distorted LiDAR Simulation (MDLS) dataset introduced in~\cite{yoon2018mapless}. This dataset represents the synthetic domain $S$ of our CycleGAN-based DA approach discussed in Section~\ref{cycGAN}. The MLDS dataset was generated from high fidelity simulated urban traffic environments from the CARLA simulator~\cite{Dosovitskiy17} using a simulated Velodyne HDL-64E sensor. The dataset is originally meant for studying the effect of the motion distortion resulted from a moving vehicle-based 3D LIDAR sensor on the generated point cloud data. The dataset consists of two sequences of point cloud data from urban traffic environment involving between 60 to 90 moving vehicle, each one with an average duration of five minutes which results in total 6K point cloud scans. The dataset was annotated with the position of the vehicles in the scene. For our DA task, we first preprocessed the point cloud scans in order to get a BEV image of each scan according to the method introduced in~\cite{li20173d}. As a result, we get a total of 6K BEV point cloud images similar to the right image shown in Fig.~\ref{fig:syn_velo}. The second dataset we utilised for the real domain $R$ of our CycleGAN-based DA approach is the BEV benchmark data from the KITTI dataset~\cite{Geiger2012CVPR}. The BEV benchmark data consists of 7481 training images and point cloud scans and 7518 test images and point cloud scans. The point cloud data was captured using a real 3D LiDAR sensor the Velodyne HDL-64E sensor. The dataset contains annotations for multiple objects in the traffic scene such as vehicles, pedestrians and cyclists. Similar to the pre-processing step we have done for the MLDS dataset we did it as well for the KITTI dataset in order to get BEV point cloud images like the one shown on the left in Fig.~\ref{fig:syn_velo}. In our experiments for training our CycleGAN-based DA approach, we used a total 6K BEV point cloud images from the MLDS dataset and the 7481 BEV point cloud images of the training split from the KITTI dataset.
 \begin{figure}[t]
 \centering
	\includegraphics[width=\columnwidth,height=13cm, keepaspectratio=true]{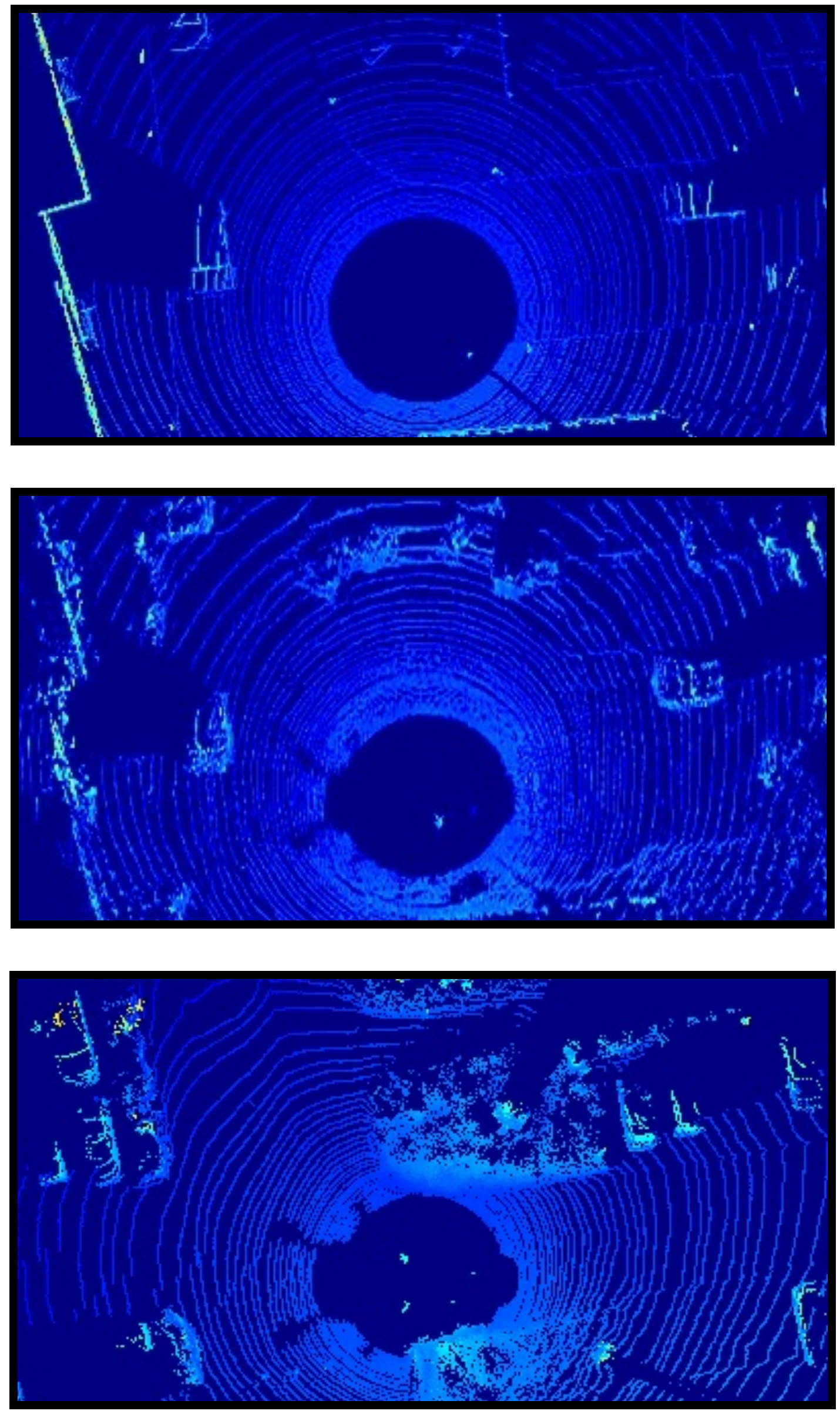}
		\caption{Qualitative results for the proposed CycleGAN-based method for DA between synthetic and real BEV point cloud data. The first row is the input the synthetic BEV point cloud image from~\cite{yoon2018mapless}. Second row is the transformed real BEV point cloud image using the proposed method. Third row is the correlated real BEV point cloud image from the KITTI dataset~\cite{Geiger2012CVPR}.} 
	\label{fig:da_velo}
\end{figure}
\smallbreak
Similarly, for the task of the vehicle detection from BEV point cloud images we used the same aforementioned two datasets (MLDS and KITTI) in addition to the domain adapted BEV images from synthetic to real for training our YOLOv3 model. Since our ultimate goal in the vehicle detection task is to identify vehicles in real BEV point cloud images. Thus, we further split the total 7481 real BEV images from the KITTI dataset into 4K for training our YOLOv3 model and 3481 for testing the model. 

	\begin{figure*}[t]
		\centering
		\includegraphics[height=7cm, keepaspectratio=true]{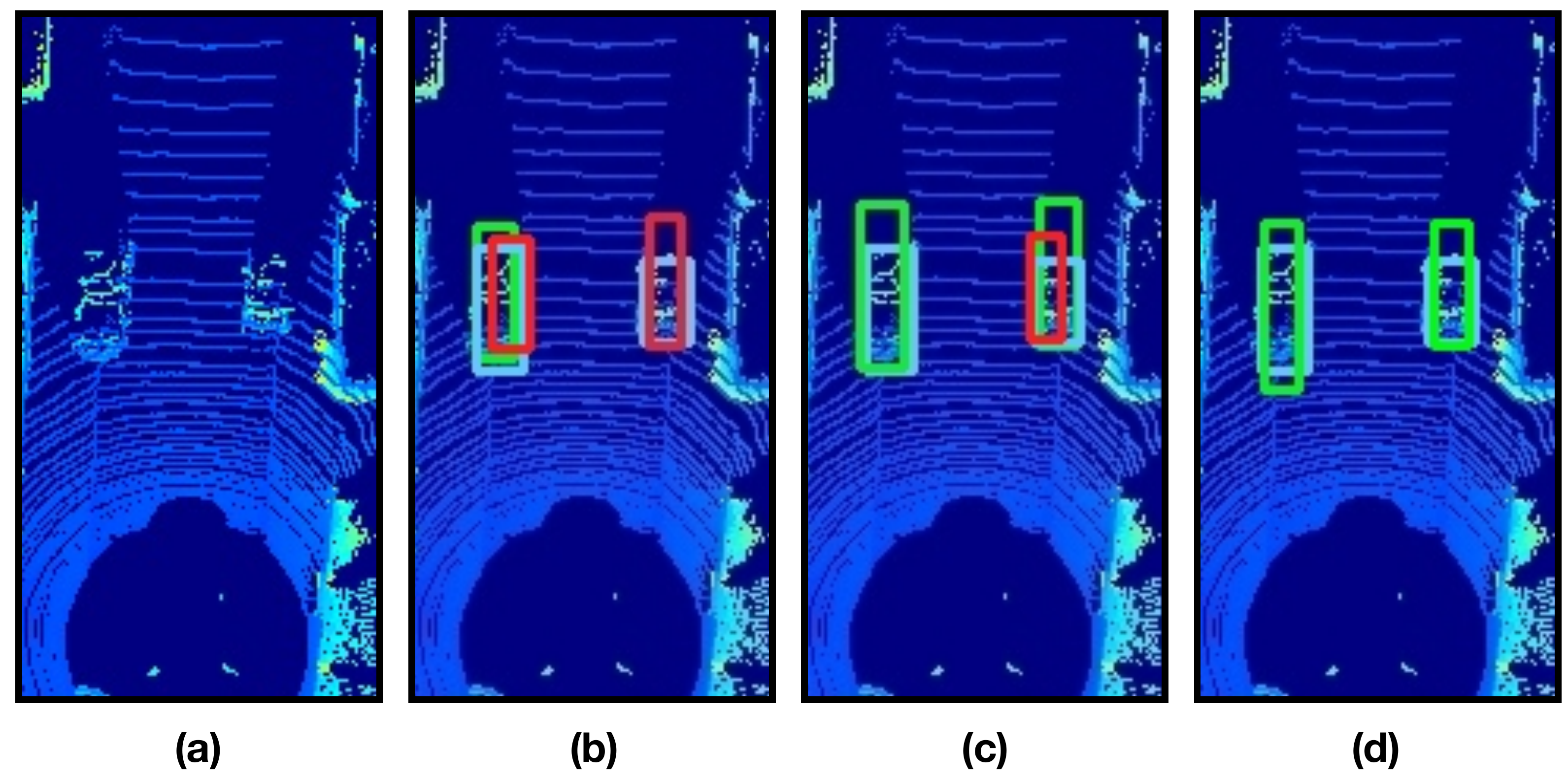}
		\caption{Qualitative results on the KITTI BEV point cloud dataset for the vehicle detection task. From left to right, a) the input BEV image , b) bounding box detections from $YOLO_K$ model, c) bounding box detections from $YOLO_{KS}$ model, d) bounding box detections from $YOLO_{KR}$ model.}
		\label{fig:frmwrk}
	\end{figure*}
\subsection{Results and Discussion}
Firstly, in order to evaluate the effectiveness of our proposed CycleGAN based DA approach for the vehicle detection task from real BEV point cloud images. In fig.~\ref{fig:da_velo}, we show qualitative results of the trained CycleGAN-based DA approach between synthetic and real BEV point cloud images. In the first row of the figure is the input synthetic BEV point cloud image to our model. The second row represents the output from the generator $G_{S\rightarrow R}$ of our trained CycleGAN model. The third row shows one sample of a real BEV point cloud image from the KITTI dataset. As it can be noticed, the generated BEV point cloud from our CycleGAN model is mimicking and trying to be consistent with the same structure exist in the real BEV point cloud image from KITTI. More specifically, the generated image captures pretty well the structure of the vehicles and the distortion/noise artefacts from resulting from the real Velodyne 3D LiDAR sensor. 

\smallbreak
For having more quantitative evaluation of our proposed CycleGAN based DA approach for the vehicle detection task, we trained two YOLOv3 models, the first one $YOLO_S$ is trained using the 6K synthetic BEV point cloud images, while the other one $YOLO_R$ is trained using the same 6K BEV point cloud images but the DA versions of them after feeding them to our trained CycleGAN model and getting its predicted DA real BEV point cloud images. Furthermore, we trained three additional YOLOv3 models with the only difference in the type of training data. The first model $YOLO_K$ which as the name implies is trained on the 4K training split BEV point cloud images from the KITTI dataset. The second model $YOLO_{KS}$ is trained using on the 4K images from the KITTI dataset with an additional 6K synthetic BEV point cloud image from the MLDS dataset. The third and final model $YOLO_{KR}$ is trained using the same amount of data to the $YOLO_{KS}$ model, however instead of the MLDS synthetic BEV images we used the DA version predicted from our CycleGAN model.

\begin{table}[t]
	\caption{Comparison between our 5 trained YOLOv3 models on the same testing split BEV point cloud images from the KITTI dataset~\cite{Geiger2012CVPR}. Higher is better.}
	\label{tab:pr}
	\centering
	\begin{tabular}{c|c|c}
		\toprule
		Model & Training Data & Average Precision (AP)\% \\
		\toprule  
		$YOLO_S$ & SYN (only) & 29.93\\
		$YOLO_R$ & DA (only) & \textbf{34.78}\\
		\midrule 
		$YOLO_K$ & KITTI (only)& 57.26\\
		$YOLO_{KS}$& KITTI+SYN & 59.16\\
		$YOLO_{KR}$& KITTI+DA & \textbf{64.29} \\
		\bottomrule
	\end{tabular}
\end{table}	

\smallbreak
In Table~\ref{tab:pr}, we report the performance of the total 5 YOLOv3 models we mentioned earlier when all are tested on the same 3481 testing real BEV point cloud images from the KITTI dataset. The evaluation metric we used is the average precision score (AP) which summarises the precision-recall curve that commonly used for evaluating object detectors. As it can be noticed from the table, the $YOLO_R$ model outperformed the $YOLO_S$ with more than 4\% in AP score which proves our claim that our CycleGAN-based DA approach for the BEV point cloud images are more efficient than pure synthetic ones for the vehicle detection task. Additionally, the best performing model with 64.29\% in AP score is the $YOLO_{KR}$, which again proves the benefits of using domain adapted BEV point cloud images over the purse synthetic ones. This prevalent from Table~\ref{tab:pr} by the low AP scores from the $YOLO_K$ and the $YOLO_{KS}$ models which achieved only AP score of 57.26\% and 59.16\% respectively. 

\smallbreak
For a qualitative measuring of the performance of the trained YOLOv3 models, in Fig.~\ref{fig:frmwrk}, we show a) input sample BEV point cloud image, b), c) and d) the detected bounding boxes (in green colour) from models $YOLO_K$, $YOLO_{KS}$ and $YOLO_{KR}$ respectively. The ground truth annotations are highlighted in the light blue colour, while the false or miss-detected objects are highlighted in red colour. As it can be shown, our model $YOLO_{KR}$ gives an accurate detection with the lowest false-positive rate.

\section{Conclusion}\label{concl}
In this work, we have introduced a framework for domain adaptation between synthetic and real BEV point cloud images for the vehicle detection task. The proposed framework utilises deep generative adversarial networks, CycleGAN for the domain adaptation task. Then, given the domain adapted BEV point cloud images we trained a series of object detection models based on state-of-the-art deep ConvNet-based model, YOLOv3. The trained models have shown the effectiveness of the proposed DA approach for the vehicle detection task from real BEV point cloud images. Furthermore, we have evaluated the performance of the trained models on the testing split from real BEV point cloud images from the KITTI dataset. The best performing model was the one utilising our domain-adapted BEV point cloud images which achieved the highest average precision score of 64.29\% with an improvement of more than 7\% over the compared baseline approaches.

\section*{Acknowledgement}
This research was fully supported by the Institute for Intelligent Systems Research and Innovation (IISRI) at Deakin University.

	\bibliographystyle{IEEEtran}
	\bibliography{IEEEabrv,Ref}

\end{document}